\documentclass[11pt]{article}

\usepackage[final]{acl}

\usepackage{times}
\usepackage{latexsym}

\usepackage[T1]{fontenc}

\usepackage[utf8]{inputenc}

\usepackage{microtype}

\usepackage{inconsolata}

\usepackage{graphicx}

\usepackage{amsmath}
\usepackage{amssymb}
\usepackage{booktabs}
\usepackage{multirow}
\usepackage[table]{xcolor}
\usepackage{float}
\usepackage[T1]{fontenc}
\usepackage[T5]{fontenc}

\title{On the Optimality of Kinship Naming: an Information-theoretic Approach}

\author {
    Phong Le\textsuperscript{\rm 1},
    Mees Lindeman\textsuperscript{\rm 2},
    Raquel G. Alhama\textsuperscript{\rm 2} \\
    \textsuperscript{\rm 1}School of Computer Science; University of St Andrews\\
    \textsuperscript{\rm 2}Institute for Logic, Language and Computation; University of Amsterdam.\\
    pl200@st-andrews.ac.uk, mees@lindeman.nu, 
    rgalhama@uva.nl
}

\begin{document}
\maketitle
\begin{abstract}
The structure of naming systems in natural languages hinges on a trade-off between high informativeness and low complexity. Focusing on the domain of kinship naming, we analyze such trade-off while addressing simplifying assumptions of prior work, namely: (i) universal communicative need across languages, and (ii) optimal listeners. To that aim, we collect data from four different languages, and analyze how different communicative needs and variations in the listener
model influence the informativeness–complexity
trade-off. Adopting a referential game setup from emergent communication, we further show that trade-off optimality is not only theoretically achievable but also emerges empirically in learned communication systems. 
\end{abstract}
\section{Introduction}

Languages vary in how they structure meaning partitions. 
For instance, unlike in English, Vietnamese offers different words for \emph{aunt}, depending on whether she is the younger or elder sister of one's mother or father. A body of work suggests that languages do not explore the space of possible semantic partitions freely, as evidenced by constrained and recurrent cross-linguistic patterns~\citep{Kemp2012, regier2015word, Zaslavsky2018, Kemp2018, Carr2020, Chaabouni2021CommunicatingSystems}. Instead, the structure of object naming systems appears to reflect pressures for communicative efficiency and learning. 
In particular, these studies suggest that languages tend to evolve toward object naming systems that approximate an (often near-)optimal trade-off between informativeness and complexity. 

This trade-off can be formalized with constructs from information theory~\cite{shannon1948mathematical}. Informativeness is typically quantified by the amount of information preserved in communication, while complexity measures how concisely a language compresses meaning into words. Using these notions, frameworks define optimal trade-off boundaries: curves along which no system can reduce complexity without increasing information loss.

On the domain of kinship, \citet{Kemp2012} measure complexity based on shortest semantic description length, and demonstrate that kinship systems lie near an optimal trade-off frontier. However, such descriptions are not grounded on empirical data, and a universal communicative need is assumed across languages--even though cultural differences can be reflected in kinship communicative need \citep{AnandRegier2023}. A relevant limitation of their framework is that their measure of complexity is not analytically tractable to derive a closed-form expression for the trade-off curve. 
In contrast, the Information Bottleneck (IB) framework ~\citep{tishby2000information} allows for the derivation of theoretical, closed-form \emph{approximations} of the optimal trade-off frontier. 

Here, we apply the IB framework to the case of kinship. Our contributions are as follows. First, inspired by multiclass classification, we adapt the information loss in the IB framework. This allows us to (i) apply the framework to discrete domains, and (ii) deduce the optimal trade-off curve directly from it--which prevents the need to add additional parameters to estimate it empirically, avoiding abnormal situations for some values of those parameters. Second, we present a novel dataset which estimates communicative need in four languages, and we use it to analyze how different assumptions (such as different communicative needs or variations in the listener model) influence the informativeness--complexity trade-off in communication. Third, adopting a referential game setup commonly used in emergent communication~\citep{Lazaridou2020,Lazaridou2017Multi-AgentLanguage,Havrylov2017EmergenceSymbols,chaabouni-emergent2022}, we show that our notion of optimality is not only theoretically achievable but also emerges empirically in learned communication systems. 

\section{Background}


\subsection{Kinship Naming}

Every society uses language to refer to family members, or \emph{kin}, through a system of lexical items that categorize familial roles (e.g., \texttt{father} and \texttt{sister}). A relevant source of cross-linguistic variation lies in how kinship meaning space is partitioned, or in other words, which family members are considered part of the same semantic category. For example, in English, both maternal and paternal grandmothers fall under the same category (\texttt{grandmother}). In contrast, Vietnamese differentiates both lineage and age, employing distinct terms for maternal versus paternal grandparents, and for older versus younger siblings~\cite{van1989vietnamese}. 

\citet{Kemp2012} observe that kinship systems appear to reflect a trade-off between informativeness---here, the ability to distinguish between kin members based on kinship names---and complexity. The latter relies on a symbolic rule system to characterize the meaning of kinship names through logical compositions of primitives (e.g., \texttt{mother} would be \emph{PARENT} \& \emph{OLDER} \& \emph{FEMALE}). Complexity is then quantified as the minimal number of logical rules needed to generate the system. Informativeness is measured as the expected Kullback-Leibler (KL) divergence between intended and inferred referents, averaged over a communicative need distribution.


\subsection{Information-Bottleneck framework}


\citet{Zaslavsky2018} introduce an information-theoretic framework for quantifying the trade-off between informativeness and complexity in lexical systems. Focusing on the continuous domain of color, they propose that  color naming systems approximate optimal trade-offs by compressing perceptual meanings into words in a manner consistent with the Information Bottleneck (IB) principle~\citep{tishby2000information}, provided that listeners are optimal in the Bayesian sense. 

Concretely, in the IB framework for color naming, 
a Speaker and Listener communicate about colors \( u \in \mathcal{U} \), where \( \mathcal{U} \) represents a continuous perceptual space. Upon perceiving a color \( u \), the Speaker selects a meaning \( m \), modeled as a distribution over \( \mathcal{U} \), and then generates a name \( w \) using encoder \( q_s(w|m) \). The Listener decodes the message using:
\[
\hat{m}_w(u) = \sum_m \tilde{q}_s(m|w) \, m(u),
\]
which defines the \emph{Bayesian-optimal listener}.

Assuming a Bayesian-optimal listener, complexity is quantified as mutual information between the Speaker's meaning variable \( M \) and the word variable \( W \):
\[
I_{q_s}(M; W) = \sum_{m, w} p_s(m) \, q_s(w|m) \log \frac{q_s(w|m)}{p_s(w)},
\]
while informativeness is captured by the mutual information \( I_{q_s}(W; U) \), measuring how much information the word conveys about the original object.

Following the IB principle, an optimal trade-off between complexity and informativeness is achieved by minimizing the following objective:
\[
\mathcal{F}_\beta[q_s(w|m)] = I_{q_s}(M; W) - \beta I_{q_s}(W; U),
\]
where \( \beta \ge 1 \) is a trade-off parameter that balances compression and informativeness.

Recently, the IB framework has been applied to discrete domains, such as artifact and animal naming~\cite{Zaslavsky2019} and personal pronouns~\cite{zaslavsky2021let}. In these settings, $\mathcal{U}$ denotes the set of all possible objects (e.g., animals), and the meaning $m$ represents the speaker’s uncertainty in recognizing the target. That is, it captures the probability that the speaker confuses the intended object with another similar one (e.g., a dog versus a wolf).

As a result, one can straightforwardly apply the IB framework to kinship naming. However, this approach has three main drawbacks. First, the framework is designed to capture cognitive constraints such as limited memory or confusability between similar objects. While this perspective is appealing when the space of objects is large (e.g., an effectively infinite color spectrum), it is unclear how to estimate and formally encode these cognitive limitations in the object--representation mapping. Existing instantiations~\cite{zaslavsky2021let} often rely on \textit{ad hoc} choices. Second, the informativeness term $I_{q_s}(W;U)$ does not characterize the full communication process, as it leaves out the impact of the listener. In practice, the IB framework typically assumes a Bayesian-optimal listener and a noiseless communication channel, assumptions that are not always realistic. For instance, when the listener has a hearing impairment or when the signal is corrupted by channel noise. Finally, the IB framework does not account for the influence of the communicative need distribution.

\section{Framework}
\label{sec:ib}

We propose an adaptation of the IB framework to the problem of kinship naming that addresses the three issues pointed out above. \footnote{Following \citet{Kemp2012,Zaslavsky2018} and many others on the emergence of language and communication, we do not assume that agents engage in pragmatic reasoning. In particular, speakers do not choose utterances by reasoning about how listeners will interpret them, nor do listeners reason about why a rational speaker would have selected a particular utterance, as in the Rational Speech Acts (RSA) framework \cite{frank2012predicting}. Nevertheless, empirical evidence suggests that incorporating RSA-style pragmatic reasoning can potentially improve the linguistic efficiency of language use when contextual information is available \cite{kobrock2026role}.}
First of all, for domains with only a small number of discrete objects, such as kinship, it is difficult to justify the assumption that a human speaker cannot reliably distinguish between two items (e.g., two family members), as would be the case in color naming due to its continuous nature. In such cases, $M$ and $U$ effectively become identical random variables. Therefore, we eliminate $M$, and consider the following scenario. 

There is a pool $\mathcal{U}$ of kinship relations, associated with a communicative need distribution $p(\cdot)$. A Speaker aims to communicate about a kinship relation $u \sim p(u)$ by selecting a message (or name) $w \in \mathcal{W}$ with encoding probability $q_s(w | u)$. The Speaker sends this message to a Listener, who then attempts to infer the intended kinship using a decoder $q_l(u | w)$.

\begin{figure}[h]
    \centering
    \includegraphics[trim=0mm 0mm 0mm 0mm, clip, width=1.0\linewidth]{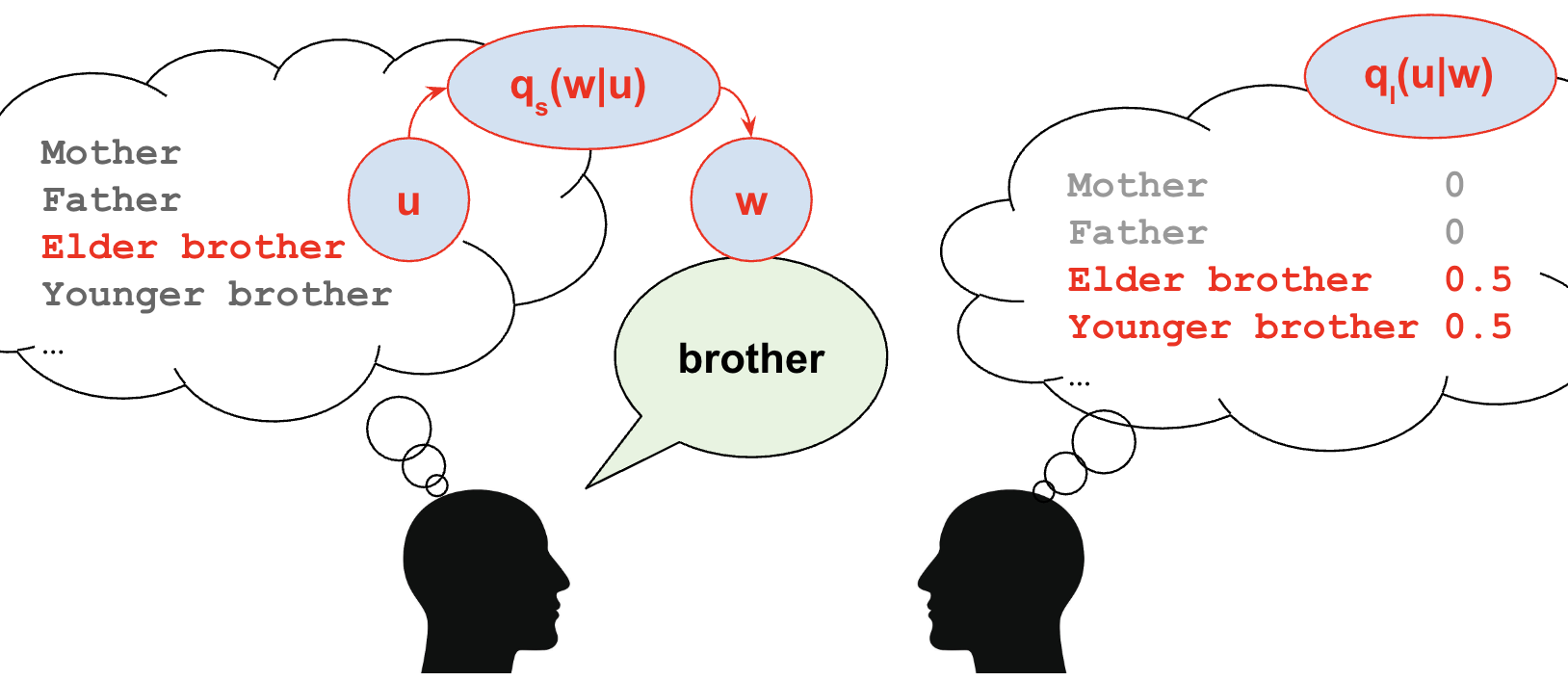}
    \caption{Illustration of two English-speaking agents playing the kinship naming game. The Speaker (left) selects a family member and produces a name. The Listener (right) receives the name and infers which member is being referred to.}
    \label{fig:kinship_game}
\end{figure}

As illustrated in Figure~\ref{fig:kinship_game}, consider an English-speaking Speaker who wishes to refer to $u = \texttt{elder brother}$. Due to the structure of English kinship terminology, the most specific term available is $w = \text{``brother''}$, and the Speaker deterministically selects it, i.e., $q_s(\text{``brother''} | \texttt{elder brother}) = 1$. Upon receiving the message, an English-speaking Listener infers that the referent could be either \texttt{elder brother} or \texttt{younger brother}, assigning equal probabilities: $q_l(\texttt{elder brother} | \text{``brother''}) = q_l(\texttt{younger brother} | \text{``brother''}) = 0.5$.

\subsection{Complexity}
\label{subsection:complexity}
Like in the IB framework, we quantify complexity as the amount of information the Speaker compresses through its encoder $q_s(w|u)$. This is measured by the mutual information between the object random variable $U$ and the message random variable $W$:
\begin{equation}
    \label{eqn:complexity}
    C = I_{q_s}(U;W) = \sum_{u, w} p(u)\, q_s(w|u) \log \frac{q_s(w|u)}{p_s(w)}
\end{equation}
where $p_s(w) = \sum_{u \in \mathcal{U}} p(u)\, q_s(w|u)$ is the marginal distribution over messages. Intuitively, this quantity captures the average amount of information that a message $w$ conveys about the intended object $u$. If the same message is used for every meaning, mutual information is zero. Conversely, if each meaning is encoded with a distinct message, mutual information reaches its maximum value---the entropy $H(U)$. Thus, complexity is bounded by $0 \le C \le H(U)$.


\subsection{Information Loss}
Different from the IB framework, we define information loss to quantify how much information is \emph{not} preserved \emph{throughout the whole communication process}, and is therefore directly related to the errors made by the Listener when picking a referent. We define information loss as the expected cross-entropy between the true referent and the Listener’s prediction, i.e., the standard loss function in multi-class classification:
\begin{equation*}
    \label{eqn:infoloss}
    L = -\mathbb{E}_{u \sim p}\, \mathbb{E}_{w \sim q_s(\cdot|u)} \log q_l(u|w).
\end{equation*}
\noindent Intuitively, the more confident the Listener can be about the intended object $u$ given the message $w$, the lower the information loss. Conversely, uncertainty in decoding leads to higher loss.

\subsection{Optimality}
\label{subsection:optimality}
Let $\tilde{q}_s(u|w) \propto q_s(w|u)\, p(u)$ be the Speaker's Bayesian decoder. We prove in Appendix~\ref{appendix:infoloss_lower_bound} that:
\begin{equation}
    L = H(U) - C + \mathbb{E}_{w \sim p_s} \left[ D_{\text{KL}}(\tilde{q}_s \Vert q_l) \right]
      \ge H(U) - C 
    \label{eqn:infoloss_lower_bound}    
\end{equation}

A direct result from this identity is that equality holds if and only if $q_l = \tilde{q}_s$, which is also entailed by the IB framework. This implies that, the optimal trade-off is achieved when the Listener's decoder exactly matches the Bayesian decoder of the Speaker. However, it is essential to emphasize that although systems with Bayesian listeners are optimal in this formal sense, they are not necessarily equally effective. Human communication succeeds because it balances informativeness and simplicity, often sacrificing some simplicity to convey sufficient meaning. In contrast, low-information systems can also be considered optimal—due to their high simplicity—but are practically unhelpful because they are too ambiguous. Thus, inline with previous work e.g. \citet{Kemp2012,Zaslavsky2018}, optimality in this framework must be understood in the context of both information loss and complexity.

Interestingly, Equation~\ref{eqn:infoloss_lower_bound} allows us to measure Euclidean distance from a given communication system to the curve \emph{analytically} by
\(
d = \frac{L-H(U)+C}{\sqrt{2}} 
\).

\subsection{Need-agnostic Theoretical Optimal Curve}
\begin{figure}
    \centering
    \includegraphics[trim=10mm 5mm 5mm 5mm, clip,width=1.0\linewidth]{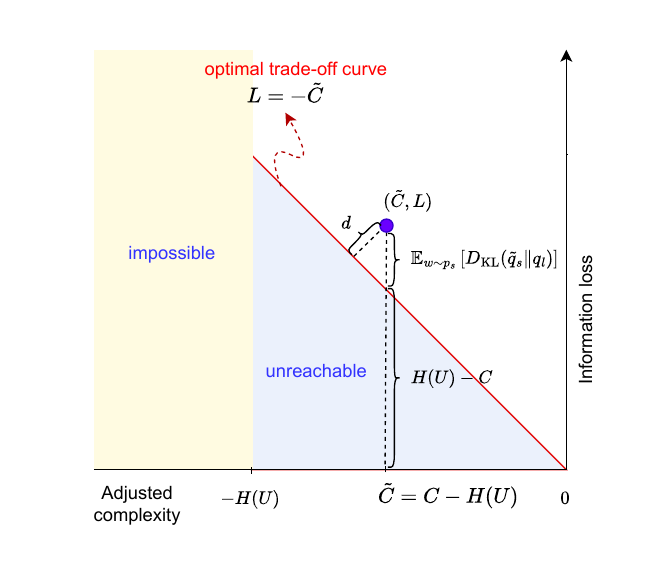}
    \caption{Visualization of the optimal trade-off curve (red line) and the feasible region (white area) encompassing all valid communication systems. 
    The Euclidean distance from a system with complexity $C$ and information loss $L$ (the blue dot) to the curve is given by $d = \mathbb{E}_{w \sim p_s} \left[ D_{\text{KL}}(\tilde{q}_s \Vert q_l) \right] / \sqrt{2}$.
    }
    \label{fig:adjusted_optimal_curve}
\end{figure}

The curve \( L = H(U) - C \) described above depends on the communicative need distribution \( p(u) \), due to the inclusion of the entropy term \( H(U) \). This dependency complicates cross-linguistic comparisons of optimality, as different need distributions induce different optimal curves in the complexity–information loss space
To address this issue, prior work—including \citet{Kemp2012} and \citet{Zaslavsky2018}—assumed a universal, fixed need distribution across languages.

Here we propose a more principled alternative: by defining an adjusted complexity \( \tilde{C} = C - H(U) \le 0 \), we transform the optimal curve into the simplified form \( L = -\tilde{C} \). This reformulation removes the dependence of the optimal trade-off curve—\emph{but not of information loss or complexity themselves}— on \( p(u) \), while preserving Euclidean distance. As a result, it enables clear visualization across languages, regardless of their underlying communicative needs. Notably, the smaller \( \tilde{C} \) is, the less complex the system is, and the more \emph{capacity} it has to increase in complexity. When \( \tilde{C} = 0 \), the system reaches its maximum allowable complexity under the given need distribution.

Figure~\ref{fig:adjusted_optimal_curve} illustrates the optimal trade-off curve, along with the feasible region in which all valid communication systems must lie. 




\section{The Emergence of Kinship Naming}
\label{section:simulation description}
In this section, we present kinship emergence from communication, based on the familial structure introduced by \citet{Kemp2012}. This structure includes 33 family members spanning five generations, with a designated \emph{ego} representing the Speaker ( Figure~\ref{fig:kintree}). Since kinship terms in some languages—such as Korean—depend on the (binarized) gender of the speaker, we consider two ego identities: Alice (female) and Bob (male). 

We examine two types of kinship naming terminologies: human (i.e., based on a sample of natural languages) and neural network-based (i.e., emerging from neural-network (NN) agents simulations). In the former case, we assume a simple probabilistic model to formalize encoding and decoding of messages by Speaker and Listener, while in the latter, the encoder and decoder are learned with the same neural network agents, which develop their own kinship terminology while playing a referential game. We refer to these systems as HP (for Human-Probabilistic) and NN, respectively.

\subsection{Human (HP) Kinship Systems}
We investigate kinship naming across four languages: English, Dutch, Spanish, and Vietnamese. We estimate the communicative need distribution $p(u)$, the Speaker's encoder $q_s(w|u)$, and the corresponding Bayesian decoder $\tilde{q}_s(u|w)$ using frequency counts extracted from text corpora, as in \citet{Kemp2012}. However, unlike this study, we estimate a separate need distribution for each language.

Concretely, for each family member $u$, we compile a set of commonly used referring expressions $T(u)$, such as, ``mother'', ``mommy'', ``mom'' for \texttt{mother}. For each term $w \in T(u)$, we estimate $\text{count}(u, w)$—the number of times $w$ refers to $u$—by searching the corpora using possessive constructions with the first person singular pronoun in each language, such as ``\emph{my} mother''. If a term is polysemous (e.g., ``brother'' can refer to either \texttt{elder brother} or \texttt{younger brother}), the count is evenly divided among all plausible referents $u$.

We then compute the total frequency of each referent $u$ by summing the counts across all its corresponding terms
\(
\text{count}(u) = \sum_{w \in T(u)} \text{count}(u, w)
\).
Let $\mathcal{F}$ be the set of all family members excluding \texttt{ego}, 
we finally estimate required distributions for speakers
\begin{equation*}
    p(u) = \frac{\text{count}(u)}{\sum_{v \in \mathcal{F}}\text{count}(v)} \; ; \;\; 
    q_s(w|u) = \frac{\text{count}(u,w)}{\text{count}(u)}
\end{equation*}
\[
    \tilde{q}_s(u|w) = \frac{q_s(w|u)p(u)}{\sum_{v \in \mathcal{F}} q_s(w|v)p(v)}
\]

Estimating the listener decoder $q_l(u \mid w)$ is inherently challenging, as text corpora rarely provide explicit information about how listeners interpret messages. To overcome this limitation, we simulate a large number of listener decoders in our experiments to systematically explore other potential trade-offs in human communication.


Further details regarding the corpora used and the counts are presented in Appendix~\ref{appendix:counts}.

\subsection{Emergent (NN) Kinship Systems}

In order to prompt the emergence of a neural-network based kinship system, we frame the model task as a referential game in which the referents are family members (see Figure~\ref{fig:kinship_game}).  The NN-Speaker is given access to the full family tree along with a randomly selected target individual. Based on this input, the Speaker generates a message intended to identify the target. Upon receiving the message and observing a candidate set that includes the target, the NN-Listener must infer which candidate the NN-Speaker is referring to.

\subsubsection{Input encoding}
\begin{figure}
    \centering
    \includegraphics[trim=10mm 1mm 1mm 0mm, clip, width=1.0\linewidth]{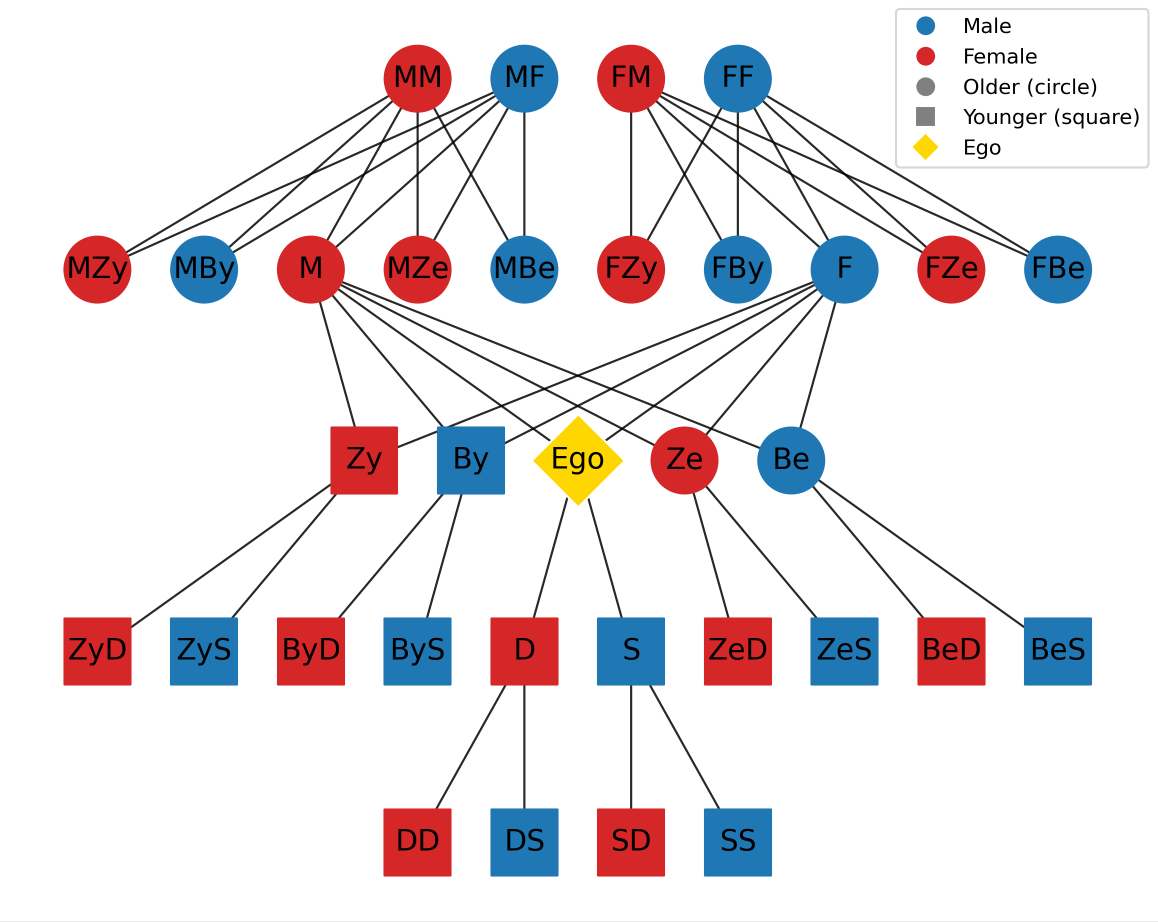}
    \caption{The kinship graph is adapted from the familial structures described by \citet{Kemp2012}. Nodes are labeled using abbreviations, where ``F'', ``M'', ``B'', ``Z'', ``S'', ``D'', ``y'', and ``e'' stand for ``father'', ``mother'', ``brother'', ``sister'', ``son'', ``daughter'', ``younger'', and ``elder'', respectively. For example, \texttt{MBe} denotes the ``mother's elder brother''.
    Each edge in the graph is bidirectional, labeled \texttt{parent-of} when traversing top-down and \texttt{child-of} when traversing bottom-up.} 
    \label{fig:kintree}
\end{figure}

The tree-like structure of a family motivates the use graphs \cite{akkerman-etal-2024-emergence}. A kinship graph consists of 33 nodes, including an \emph{ego} node, which can be either ``Bob'' (male) or ``Alice'' (female). The graph is adapted from the used family tree and is visualized in Figure~\ref{fig:kintree}. Each node represents an individual family member and is annotated with categorical features that encode key relational distinctions:
\emph{Gender} (male or female),\footnote{We consider only binary gender distinctions, consistent with those typically encoded in human kinship systems.}
\emph{Gender relative to ego} (equal or different),
\emph{Age relative to ego} (older or younger), and
\emph{Age relative to parent} (older or younger). 
All features are one-hot encoded and are designed to reflect the compositional kinship rules used in \citet{Kemp2012}.

To connect nodes (i.e., family members), we diverge from \citet{Kemp2012} by using only the two most primitive relationships—\texttt{parent-of} and \texttt{child-of}—which allow bidirectional traversal across generations. For example, the node \texttt{F} (father) connects to \texttt{Bob} (ego) via ``F is parent of Bob,'' and to \texttt{Be} (Bob’s elder brother) via ``F is parent of Be''; correspondingly, \texttt{Bob} and \texttt{Be} each connect back to \texttt{F} via \texttt{child-of} edges. More complex relationships in the kinship trees of \citet{Kemp2012}, such as \texttt{sibling-of}, must instead be inferred compositionally from the primitive relations. This design encourages agents to discover and exploit relational structure rather than rely on shortcut or explicitly labeled edges.

We observe that the kinship graph contains redundant information. For instance, determining whether two individuals are brothers can be achieved by checking whether they share either the same mother or the same father.\footnote{Following \citet{Kemp2012}, we do not consider step-siblings.} 
To eliminate such redundancy, we apply a pruning procedure that retains only the shortest paths from each node to the \texttt{ego}. This results in a best-first-search tree. 
We refer to this process as \emph{graph pruning}.

\subsubsection{Model architecture}
Our agents are implemented as two neural networks. The NN-Speaker encodes the kinship graph $G$ using a graph neural network $\text{GNN}_s$, producing node-level embeddings:
\[
[\mathbf{h}_1, \dots, \mathbf{h}_{33}] = \text{GNN}_s(G)
\]
\noindent where $\mathbf{h}_i \in \mathbb{R}^d$ is the representation of the $i$-th node. To generate a message, the Speaker concatenates the embeddings of the ego and target nodes, transforms them via a two-layer network, and applies the Gumbel-Softmax (GS) to sample a \emph{one-token} message $w$ from a fixed vocabulary $\mathcal{V}$:
\begin{align*}
    \text{score}_s(u) &= \mathbf{W}_\text{lex} \cdot \mathbf{W}_\text{hid} \cdot \text{cat}(\mathbf{h}_\text{ego}, \mathbf{h}_\text{target}) \\
    w &\sim \text{GS}(\text{score}_s(u))
\end{align*}
$\mathbf{W}_\text{hid} \in \mathbb{R}^{d_h \times 2d}$ and $\mathbf{W}_\text{lex} \in \mathbb{R}^{|\mathcal{V}| \times d_h}$ are trainable weight matrices, and $\text{score}_s(u) \in \mathbb{R}^{|\mathcal{V}|}$ denotes the unnormalized scores for tokens. Unless stated otherwise, we assume vocabulary size $|\mathcal{V}|=128$. 

The NN-Listener receives the same graph along with the sampled message $w$ and infers the target referent. It encodes the graph using another graph neural network $\text{GNN}_l$, identical in architecture to the Speaker's, and computes compatible scores between the message and family members:
\begin{align*}
    [\mathbf{v}_1, \dots, \mathbf{v}_{33}] &= \text{GNN}_l(G) \\
    \text{score}_l(w, i) &= \mathbf{e}_w^\top \cdot \mathbf{W} \cdot \mathbf{v}_i \quad \forall i \in [1, \dots, 33]
\end{align*}
where $\mathbf{v}_i \in \mathbb{R}^d$ is the embedding of node $i$, $\mathbf{e}_w \in \mathbb{R}^{d_h}$ is the embedding of token $w$, and $\mathbf{W} \in \mathbb{R}^{d_h \times d}$ is a trainable bilinear transformation. 
The Listener selects the node with the highest score as its prediction. In all experiments, we use three shared-parameter graph neural layers, set $d = 80$ and $d_h = 20$, and the NN-Speaker and NN-Listener do not share parameters.

Because kinship structures are inherently relational, we encode the input with Relational Graph Convolutional Network (RGCN)~\citep{schlichtkrull2018modeling}, designed for multi-relational graphs. In Appendix~\ref{appendix:architecture selection} we report alternative GNN architectures and hyperparameter configurations.

\subsubsection{Training}
From the kinship graph described above, we generate a dataset of 10,000 data points, each corresponding to a single game turn. Each data point consists of:  
(i) the full pruned kinship graph with the \texttt{ego} node uniformly sampled from $\{\text{Bob}, \text{Alice}\}$;  
(ii) a target node $u$ uniformly selected from the remaining $32$ nodes;  
(iii) a distractor set $D$ of five nodes, uniformly sampled from the remaining nodes (i.e., excluding both the \texttt{ego} and the target node).

We split the dataset into 80\% for training and 20\% for validation. The two agents are trained jointly to minimize the following loss:
\[
L = -\sum_{(u, D)} \log \frac{\exp(\text{score}_l(w, u))}{\sum_{v \in \{u\} \cup D} \exp(\text{score}_l(w, v))}
\]
where $w$ is the message generated by the Speaker for target $u$.\footnote{During training, $w$ sampled from  Gumbel-Softmax is a distribution over $\mathcal{V}$ (while a discrete token in evaluation).}
We use the Adam optimizer with a learning rate of $1 \times 10^{-3}$, training for 500 epochs using mini-batches of size 50.

\subsubsection{Evaluation}
To assess the generalizability of the learned communication protocol, we evaluate agents under a stricter criterion designed for faithful comparison with human-produced (HP) kinship systems. As in \citet{Kemp2012}, the NN-Listener must identify the target node from among \emph{all} 32 possible family members (rather than the 6-candidate context used during training). The evaluation set thus consists of $2 \times 32 = 64$ data points, covering each \texttt{ego} $\in \{\text{Bob}, \text{Alice}\}$ paired with all 32 non-ego members.

Unlike during training, the NN-Speaker operates deterministically at evaluation time, producing the most likely token: $w^* = \arg\max_{w} \text{score}_s(u)[w]$; hence $q_s(w|u)=[[w=w^*]]$. The NN-Listener computes a probability distribution over all candidates using a softmax over the score function: $q_l(u|w) \propto \exp(\text{score}_l(w, u))$.


\section{Experiments}
We conduct computational experiments to validate our theoretical framework on the optimality of kinship naming across four languages: English (en), Dutch (nl), Spanish (es), and Vietnamese (vi).

\subsection{Systems} 

To empirically evaluate the prediction that the semantic system exhibits an optimal framework if and only if the decoder is Bayesian, we need to simulate a large number of listener decoders. For each human language, we evaluate kinship naming between an HP-Speaker and one of the following types of HP-Listeners:
\begin{itemize}
    \item \emph{Bayesian} HP-Listener: uses a decoder identical to the Bayesian decoder employed by the HP-Speaker. This type is employed in the work of ~\citet{Kemp2012, Zaslavsky2018}.
    \item \emph{Non-Bayesian} HP-Listener: uses a decoder that deviates from the Bayesian decoder by randomly changing its decision with a flip rate of \(r_e \in \{0.1\%, 0.5\%, 1\%, 2\%\}\). 
    \footnote{It is important to note that the definition of these models is independent of the properties of the language. In other words, even though the non-Bayesian model is expected to perform worse than the Bayesian model in terms of decoding (or communicative success), this formulation is independent of the definitions of complexity and information loss, hence also of the extent to which their trade-off approaches optimality.}
\end{itemize}

Each Non-Bayesian condition is simulated with a population of 10 million listeners per language, resulting in a total of 40 million simulated listeners across all languages, along with four Bayesian listeners. Leveraging this large-scale simulation—and grounded in the well-established information-theoretic concepts of information loss and complexity—we assess whether communication with Bayesian listeners is indeed optimal, as predicted in Section~\ref{subsection:optimality}.

We also evaluate the communication performance of our neural network agents (NN) under each language-specific communicative need distribution.\footnote{
We use the EGG framework~\cite{Kharitonov2019EGG:Games} and the PyTorch Geometric library~\cite{fey2019fast}. 
Our source code is available at \url{https://github.com/meeslindeman/kinship}.}
The communication setup is ego-specific, with \texttt{ego} $\in \{\text{Bob}, \text{Alice}\}$. Following the configuration described in Section~\ref{section:simulation description} and summarized in Appendix~\ref{appx:hyper-parameters}, we repeat training and evaluation 50 times to account for variability across runs.

\subsection{Metrics} 
Since our primary interest lies in the optimality of the complexity-information loss tradeoff, our primary performance metric is the Euclidean \emph{distance} to the optimal curve, defined in Section~\ref{subsection:optimality}.

In addition, under each language’s communicative need distribution, we evaluate \emph{accuracy}, defined as the expected probability that a Listener correctly identifies the intended referent family member:
\(
    \mathbb{E}_{u \sim p,\, w \sim q_s(\cdot|u)} \left[ q_l(u|w) \right]
\).
This is a \emph{relaxed} variant of traditional accuracy, since the precise predictions of (natural language) Listeners are not directly observable due to polysemy (e.g., the term “brother” may refer to either \texttt{elder brother} or \texttt{younger brother}).

\begin{figure}[t]
    \centering
    \includegraphics[trim=2mm 2mm 2mm 2mm, clip, width=1\linewidth]{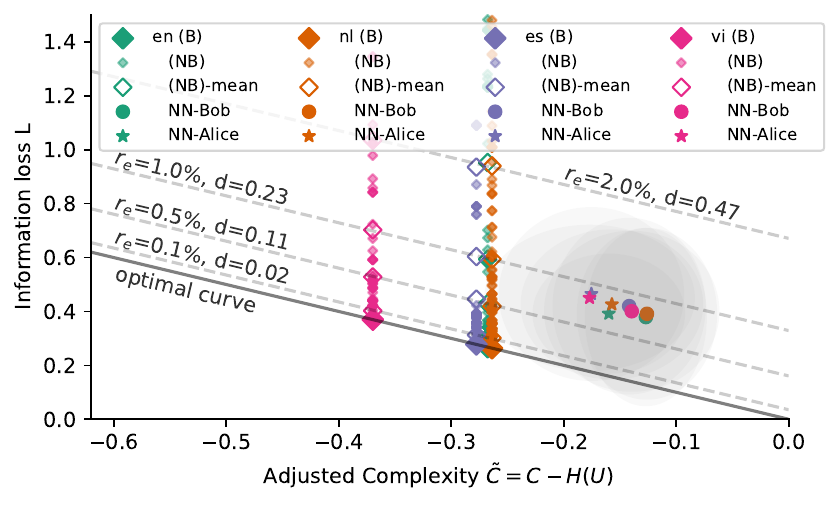}
    \includegraphics[width=1.0\linewidth]{\detokenize{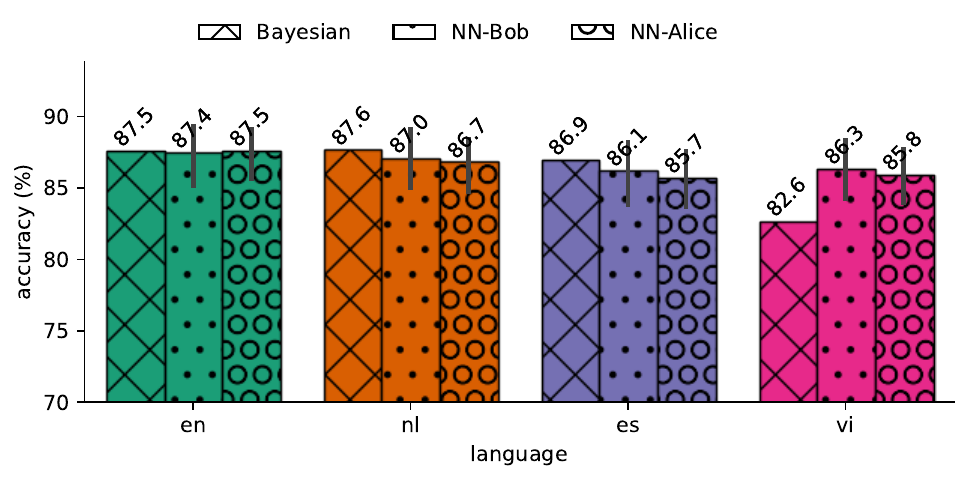}}
    \caption{
        (Top) Trade-offs for HP systems with Bayesian (B) and Non-Bayesian (NB) listeners, and neural network (NN) systems (averaged over 50 runs; standard deviations shown as light gray ellipses). Trade-offs under Non-Bayesian HP-Listener conditions collectively (denoted as NB-mean) form lines that are approximately parallel to the optimal curve. Each line is annotated with the flip rate \( r_e \) and its distance \( d \) to the optimal curve. For visual clarity, only 100 randomly selected Non-Bayesian listeners are shown per language.    
        (Bottom) Accuracy of NN Bayesian HP.
    }
        
    \label{fig:results}
\end{figure}

\subsection{Results}
\paragraph{The impact of non-Bayesian HP-Listeners.}
We evaluate four non-Bayesian HP-Listeners with error rates \(r_e\in\{0.1\%, 0.5\%, 1\%, 2\%\} \). Figure~\ref{fig:results}-top shows the trade-off under each of these HP-Listener conditions, plus the condition without noise (Bayesian HP-Listener). The figure reveals that, across all languages, communication with the \emph{Bayesian} HP-Listener lies exactly on the optimal curve (i.e., zero distance), while communication with non-Bayesian Listeners deviates progressively further from it. Moreover, HP-Listeners with lower error rates consistently yield more favorable trade-offs than those with higher error rates. These findings align with the theoretical prediction outlined in Section~\ref{subsection:optimality}:trade-off optimality depends on how closely the HP-Listener's decoder approximates the Bayesian decoder of the HP-Speaker.

Interestingly, we observe that human communication systems in the three Indo-European languages—English, Dutch, and Spanish—exhibit similar levels of adjusted complexity and information loss (and also accuracy as shown in Figure~\ref{fig:results}-bottom). In contrast, Vietnamese shows higher information loss but lower adjusted complexity, suggesting that it has more capacity to increase complexity in order to improve informativeness.

\paragraph{HP vs NN.}
Since the NN models are trained to minimize information loss, they are not inherently constrained by complexity. Fortunately, we find that early stopping serves as an effective mechanism for limiting complexity, therefore we select the checkpoint with accuracy closest to that of human communication (see Figure~\ref{fig:results}-bottom).

Figure~\ref{fig:results}-top shows that the NN communication system achieves trade-offs that are closer to the optimal curve than human communication with the non-Bayesian HP-Listener under $r_e=1\%$ flipping rate. This suggests that communication systems emerging from object naming tasks can approach theoretical optimality without incurring exhaustive complexity—similar to human communication.\footnote{Note that information loss is not an absolute indicator of accuracy—cf. the case of Vietnamese. This is analogous to what is commonly observed between log-likelihood and accuracy in classification tasks.}

\begin{figure}
    \centering
    \includegraphics[trim=2mm 2mm 2mm 2mm, clip, width=1.0\linewidth]{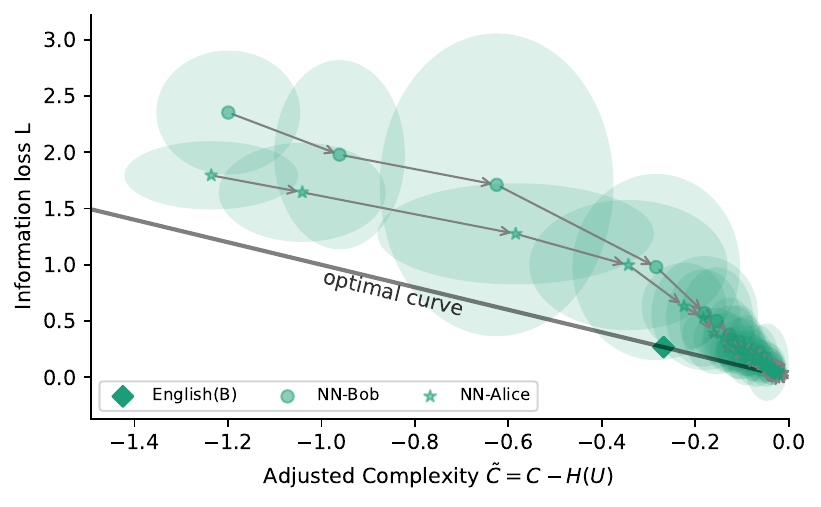}
    \caption{Trajectories of NN systems over time, recorded every 10 epochs, under the English communicative need distribution. Results are averaged over 50 runs, with standard deviation represented by ellipses.}
    \label{fig:tradeoff_evolved}
\end{figure}

\paragraph{NN naming system evolves towards optimality.}
Building on the previous results, we examine the evolution of NN communication over time. Figure~\ref{fig:tradeoff_evolved} shows the trade-offs 
under the communicative need distribution for English (results for other languages are in Appendix~\ref{appendix:evolution}). 

Initially, the emergent communication exhibits moderate complexity but high information loss, as agents rely on only a few vocabulary terms, resulting in a simple but inefficient language. As training progresses, information loss gradually decreases at the cost of increasing complexity, with the trade-off trajectory approaching the optimal curve. This trend suggests that agents progressively expand their vocabulary usage to enhance communicative success, thereby reducing information loss while incurring greater complexity.

\section{Conclusion}

We have used an information-theoretic framework for quantifying the trade-off between information loss and complexity in kinship naming systems. 
Our work accommodates non-Bayesian listeners---an important extension, as listener deviations are common in real-world settings. Indeed, even human listeners with clinically normal hearing often struggle with decoding in daily life \citep{ruggles2011}. Our work enables the estimation of trade-offs under such realistic conditions, whether in human communication or in applied scenarios where messages are subject to perturbation. Conditions for optimality can arise when the communication channel is relatively noiseless, speakers and listeners share similar linguistic biases and representations, and decoding approximates Bayesian inference.

Furthermore, we take into account the impact of cultural differences via communicative need distribution. By eliminating the dependence of the optimal trade-off curve on the need distribution, we enable clearer comparison of trade-offs across culturally shaped communication systems. For instance, Figure~\ref{fig:results} shows that trade-offs vary across language families: Vietnamese, for example, achieves lower complexity relative to its own complexity capacity compared to Indo-European languages, albeit at the cost of informativeness. The figure also suggests that optimal human communication tends to cluster within an information loss range of approximately $[0.2, 0.4]$ and an adjusted complexity range of $[-0.4, -0.2]$. A deeper analysis of these patterns, along with an expanded comparison across more languages and linguistic families, is left for future work.

Finally, we have shown that kinship naming systems emerging from NN models exhibit a trade-off comparable to that of HP systems with less than 1\% noise. This suggests that, at  least within the constraints of our NN system, optimizing only for communication accuracy (or information loss) 
and applying early stopping (accounting for limited lifespan of individuals) are sufficient mechanisms to trigger the 
evolutionary dynamics that result in the observed trade-offs.


\section*{Limitations}

\paragraph{Language complexity measurement} As in \citet{Zaslavsky2018}'s Information Bottleneck framework, we measure complexity as the mutual information between the object and word random variables. This quantity is upper-bounded by the entropy of the communicative need distribution. However, this measure does not fully capture the richness of natural languages: in principle, a language can achieve unbounded complexity by continually expanding its vocabulary.


\paragraph{Compositionality} We restrict our NN communication systems to \emph{one-token} messages, whereas natural languages often employ compositional expressions to refer to kinship relations—i.e., components of the expression systematically correspond to semantic properties of the kin category. For example, in English, the prefix \emph{grand-} consistently denotes \emph{parent-of-}, while in Spanish, the suffixes \emph{-a} and \emph{-o} typically indicate female and male family members, respectively. Such compositional strategies cannot emerge in our current setup. Additionally, because the NN-Speaker generates names deterministically, the emergent language lacks synonymy, which is common in natural languages (e.g., in English \texttt{mother} may be referred to as either ``mother'' or ``mum'').

\paragraph{Datasets} In our experiments, we compiled datasets from specific domains (see Appendix~\ref{appendix:counts}), resulting in domain-specific counts that may not be representative of all possible domains. Nevertheless, we do not consider this a fundamental limitation, as our primary aim is to demonstrate how the proposed framework can accommodate different distributions of communicative needs.

\paragraph{Ecological validity} No laboratory setting can fully reproduce the complexity of real-life language use, and our study is no exception. Nevertheless, our results show that our notion of optimality is not merely theoretically attainable, but can also emerge empirically in learned communication systems. Interestingly, at least within the constraints of our neural-network setup, optimizing solely for communicative accuracy---or, equivalently, minimizing information loss---together with early stopping as a proxy for individuals' limited lifespans, is sufficient to induce evolutionary dynamics that give rise to the observed trade-offs.

\section*{Acknowledgments}

R.G.A. was financed by the NWO SSH Open Competition XS grant (project no. 406.XS.24.01.128).
P.L. was supported by the HKUST-UStA Global Knowledge Network Awards/Joint Seed Funding.

\bibliography{custom}

\clearpage
\appendix

\section{A Lower bound for Information Loss}
\label{appendix:infoloss_lower_bound}
In this appendix we derive a lower bound for information loss. 

\begin{align*}
L =& -\mathbb{E}_{u \sim p}\mathbb{E}_{w \sim q_s(\cdot|u)}\log q_l (u|w) \\
=& -\sum_u p(u) \sum_w q_s(w|u) \log q_l(u|w) \\
=& -\sum_u p(u) \sum_w q_s(w|u) \\
&\quad \log \left[ p(u) \frac{q_s(w|u)}{p_s(w)} \cdot \frac{q_l(u|w)p_s(w)}{q_s(w|u)p(u)} \right] \\
=& -\sum_u p(u) \sum_w q_s(w|u) \log p(u) \\
& - \sum_u p(u) \sum_w q_s(w|u) \log \frac{q_s(w|u)}{p_s(w)} \\
& - \sum_u p(u) \sum_w q_s(w|u) \log \frac{q_l(u|w)p_s(w)}{q_s(w|u)p(u)} \\
=& -\sum_u p(u) \log p(u) \\
& - \sum_{u,w} p(u) q_s(w|u) \log \frac{q_s(w|u)}{p_s(w)} \\
& + \sum_w p_s(w) \sum_u \tilde{q}_s(u|w) \log \frac{\tilde{q}_s(u|w)}{q_l(u|w)} \\
=& H(U) - C + \mathbb{E}_{w \sim p_s} \left[ D_{\text{KL}}(\tilde{q}_s \Vert q_l) \right]
\end{align*}

\noindent where:
\begin{itemize}
    \item \( H(U) = -\sum_u p(u) \log p(u) \) is the entropy of the communicative need distribution;
    \item \( C \) is the complexity, as defined in Equation~\ref{eqn:complexity};
    \item \( \tilde{q}_s(u|w) = \frac{q_s(w|u)\, p(u)}{p_s(w)} \) is the Bayesian decoder of the Speaker.
\end{itemize}

Since the KL divergence is always non-negative, the information loss is lower-bounded by:
\[
L \ge H(U) - C.
\]
This bound is achieved when \( q_l = \tilde{q}_s \), i.e., when the Listener's decoder is identical to the Bayesian decoder of the Speaker.


\section{Hyper-parameters}
\label{appx:hyper-parameters}
We show relevant hyper-parameters for all experiments in Table \ref{tab:hyperparameters}.
Gumbel-softmax temperature controls the Gumbel-softmax sampling distribution: lower values tend towards a one-hot encoding, whereas higher values tend towards a uniform encoding. 

\begin{table*}[ht]
\centering
\small
\caption{Hyperparameter settings used in our experiments. The third column reports the values selected for the main study, while the last column lists the values considered during architecture search.}
\label{tab:hyperparameters}
\begin{tabular}{llcc}
\toprule
 & \textbf{Hyperparameter} & \textbf{Value (main study)} & \textbf{Values (architecture search)} \\
\midrule
\multirow{6}{*}{Model Architecture} 
    & embedding dimensions $d$ & 80 & -- \\
    & hidden dimension $d_h$ & 20 & -- \\
    & Graph neural net & RGCN & RGCN, GATv2Conv \\
    & \# graph net layers & 3 & -- \\
    & Vocabulary size $|\mathcal{V}|$ & 128 & 16, 32, 64, 128, 256 \\
    & Graph pruning & True & True, False \\
\midrule
\multirow{5}{*}{Training} 
    & Optimizer & Adam & -- \\
    & Learning rate & $1 \times 10^{-3}$ & -- \\
    & Batch size & 50 & -- \\
    & \# distractors & 5 & -- \\
    & Gumbel-softmax temperature & 1.5 & -- \\
\bottomrule
\end{tabular}
\end{table*}

\section{The evolution of NN Kinship System}
\begin{figure}[H]
    \centering
    \includegraphics[trim=2mm 2mm 2mm 2mm, clip, width=1\linewidth]{figures/trajectory_en.pdf}
    \includegraphics[trim=2mm 2mm 2mm 2mm, clip, width=1\linewidth]{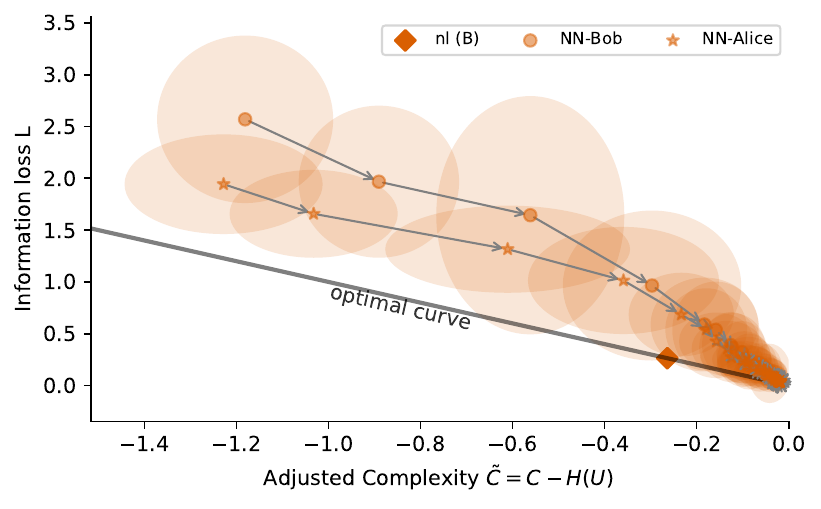}
    \includegraphics[trim=2mm 2mm 2mm 2mm, clip, width=1\linewidth]{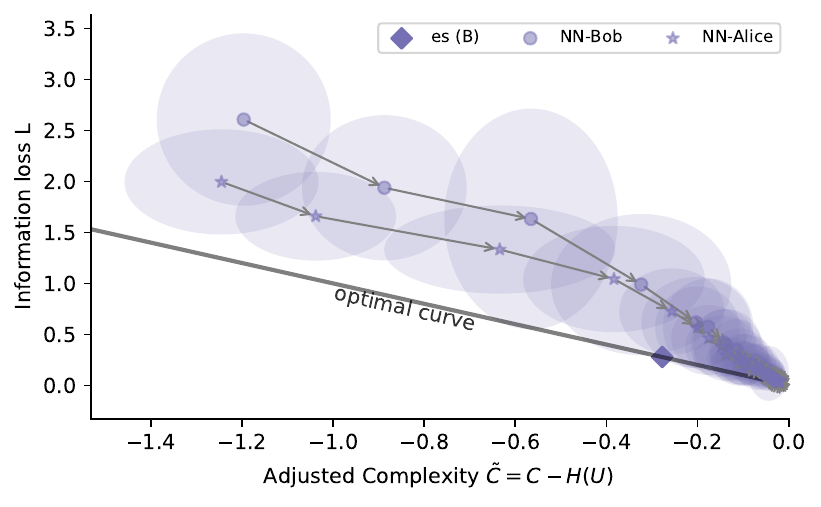}
    \includegraphics[trim=2mm 2mm 2mm 2mm, clip, width=1\linewidth]{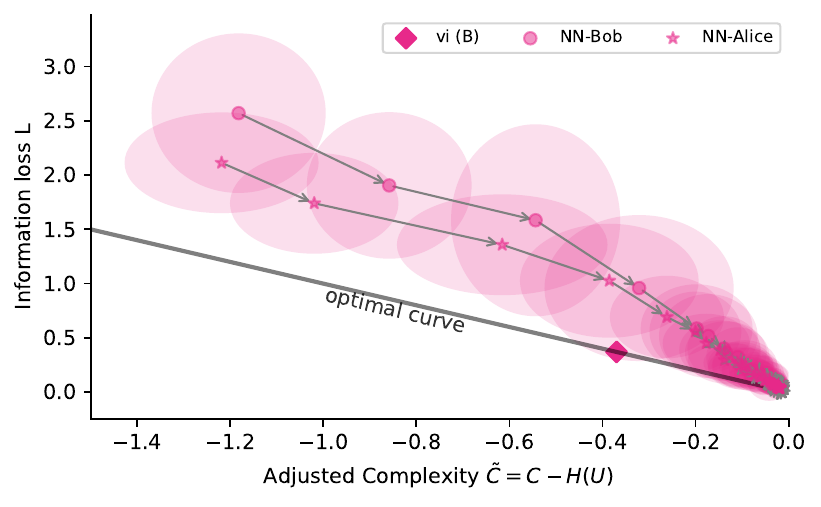}
    \caption{Evolution of complexity and information loss in ego-specific NN communication (average over 50 runs) under four communicative need distributions English, Dutch, Spanish, and Vietnamese.}
    \label{fig:all_tradeoff_evolved}
\end{figure}

We show in Figure~\ref{fig:all_tradeoff_evolved} how NN communication evolved during training in the environments of the four languages: English (en), Dutch (nl), Spanish (es), and Vietnamese (vi).

\label{appendix:evolution}


\section{Impact of Model Architecture on Performance}

\label{appendix:architecture selection}
\begin{figure}[h]
    \centering
    \includegraphics[width=1.1\linewidth]{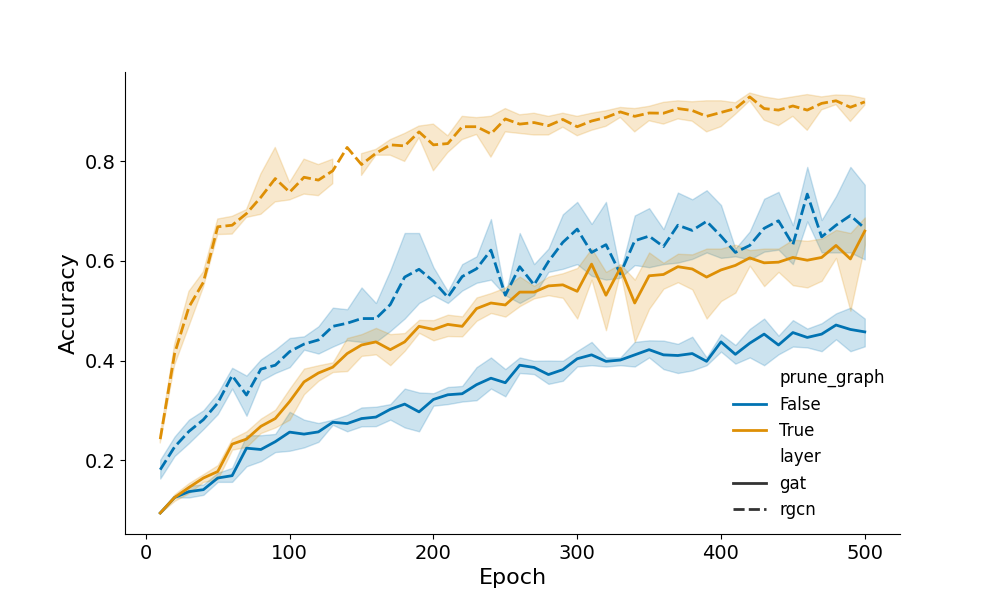}
    \caption{Evaluation accuracy, for simulations with and without pruning (n=40 runs, varying initialization).}
    \label{fig:ev_acc}
\end{figure}

\begin{figure}[h]
    \centering
    \includegraphics[width=1.1\linewidth]{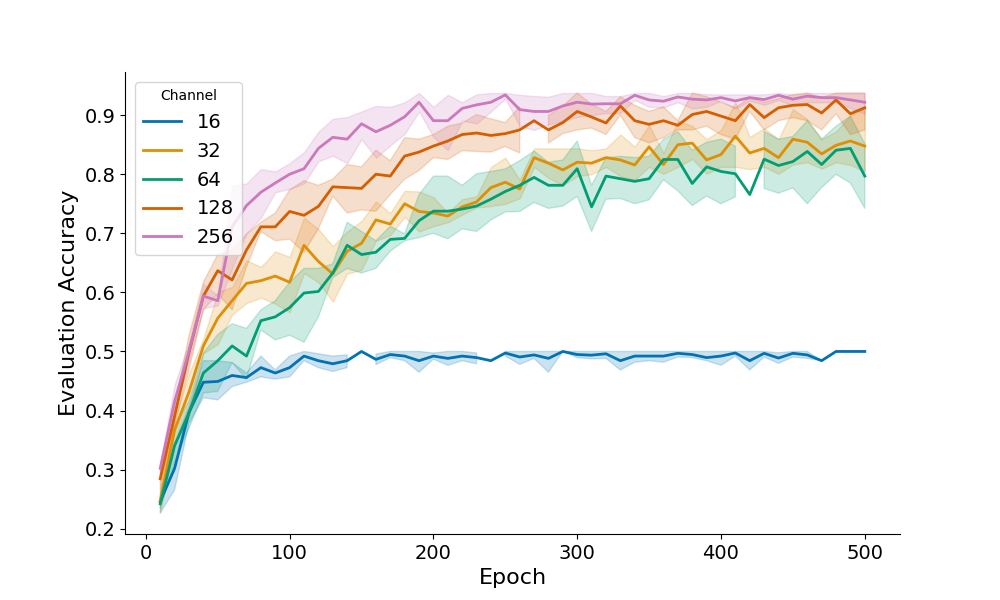}
    \caption{Evaluation accuracy, for simulations with pruning and RGCN layer (n=50 runs in total, with 10 different initializations for each vocabulary size).}
    \label{fig:voc_acc}
\end{figure}
We investigate the impact of architectural choices that led to the NN-agents used in the main study. Specifically, we examine three factors: \emph{graph pruning}, \emph{layer type}, and \emph{channel capacity} (i.e., vocabulary size) in the environment of uniform communicative need distribution.

In the first study, based on the configuration described in Section~\ref{section:simulation description} and summarized in Appendix~\ref{appx:hyper-parameters}, we construct four variants by systematically varying the use of graph pruning and the choice of layer type (either RGCN~\cite{schlichtkrull2018modeling} or GATv2Conv~\cite{brodyattentive}). As shown in Figure~\ref{fig:ev_acc}, both graph pruning and the use of RGCN layers are critical for achieving high communicative success, each contributing approximately 20 percentage points to the final communication accuracy of the NN-agents.

In the second study, we vary the vocabulary size (16, 32, 64, 128, 256) to examine the effect of channel capacity. Figure~\ref{fig:voc_acc} demonstrates that a sufficiently large vocabulary relative to the size of the object set (32 kinship terms in our case) is crucial. For instance, a vocabulary size of 16 constrains communication to approximately 50\% accuracy. In contrast, increasing the vocabulary size to 32 or greater substantially improves accuracy to 80\% and above.


\section{Human Communication with Suboptimal HP-Listeners}
\begin{figure}[h]
    \centering
    \includegraphics[width=1.\linewidth]{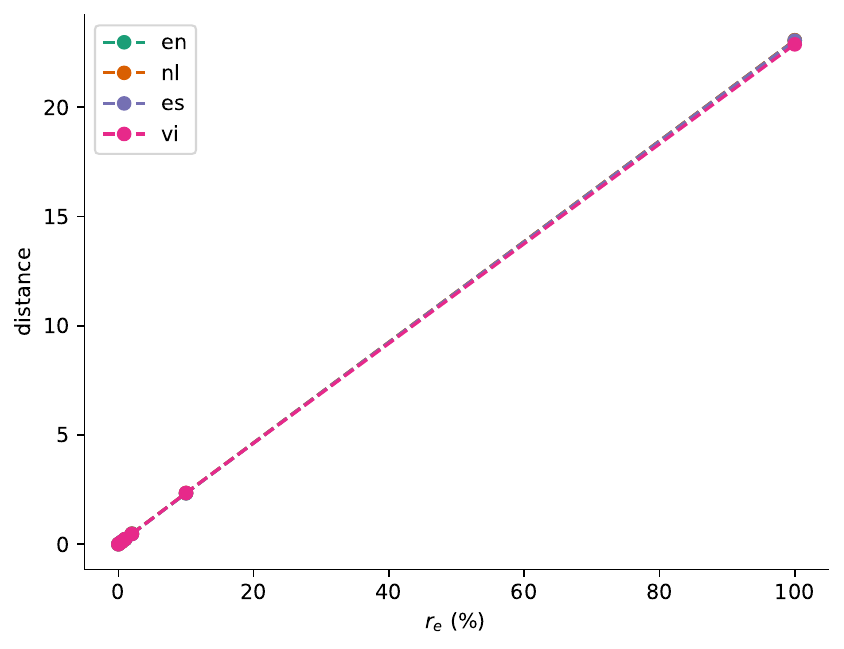}
    \includegraphics[width=1.\linewidth]{\detokenize{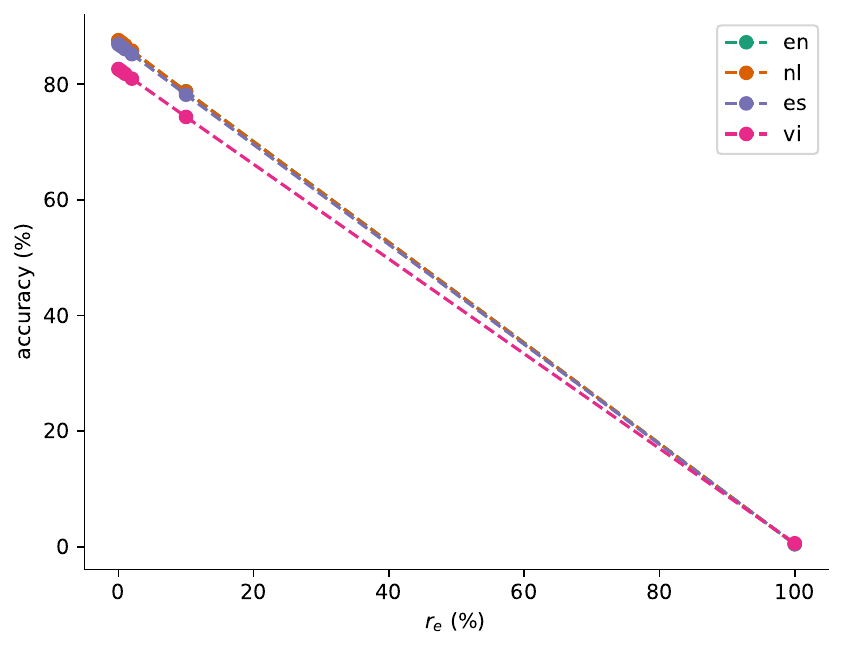}}
    \caption{(Top) Distance to the optimal curve and (Bottom) accuracy of human communication with suboptimal Listeners at varying flip rates \( r_e \in \{0, 0.001, 0.005, 0.01, 0.02, 0.1, 1.0\} \). Dashed lines indicate the linear relationship between flip rate and distance/accuracy.}
    \label{fig:human_noisy_listeners}
\end{figure}

We investigate the impact of suboptimal HP-Listeners on human communication across four language environments: English, Dutch, Spanish, and Vietnamese. Figure~\ref{fig:human_noisy_listeners} reports both the distance to the optimal trade-off curve and the accuracy of the corresponding communication systems. The results reveal a linear relationship between the HP-Listener's error rate and both distance and accuracy. This finding supports our theoretical framework, confirming that communication with non-Bayesian HP-Listeners—i.e., those that deviate more from the Bayesian Listener—results in trade-offs that lie further from the optimal curve.

\label{appendix:error_rate}


\section{Kinship counts}
\label{appendix:counts}
We extract counts of family-member and kinship-term pairs (see Table~\ref{tab:counts}) from text corpora in four languages:
\begin{itemize}
    \item \textbf{English:} We use the Corpus of Contemporary American English (COCA)~\cite{COCA}, a widely-used and balanced corpus of American English. It contains over one billion words from 1990–2019, covering eight genres such as spoken language, fiction, news, academic writing, and web content. 
    \item \textbf{Dutch:} We use the SoNaR corpus~\cite{StevinSONAR2013}, a 500-million-word reference corpus of contemporary Dutch that includes both written and spoken data. SoNaR integrates material from various sources such as newspapers, newsletters, books, websites, and transcripts.
    \item \textbf{Spanish:} We use the NOW (News on the Web) corpus from the Corpus del Español~\cite{corpusdelespanol}, which includes approximately 7.6 billion words from web-based newspapers and magazines across 21 Spanish-speaking countries, collected between 2012 and 2019. 
    \item \textbf{Vietnamese:} We use the VietVault corpus~\cite{nam_pham_2024}, a dataset filtered and curated from Common Crawl dumps prior to 2023. We sample a 5GB subset of the full  80GB of raw Vietnamese text, spanning multiple domains. 
\end{itemize}

\begin{table*}[ht]
\centering
\small
\setlength{\tabcolsep}{6pt}
\caption{Counts for (family-member, term) pairs from text corpora.}
\label{tab:counts}
\resizebox{\textwidth}{!}{%
\begin{tabular}{l|ll|cc|cc|cc|cc}
\toprule
& & & \multicolumn{2}{c}{English} & \multicolumn{2}{c}{Dutch} & \multicolumn{2}{c}{Spanish} & \multicolumn{2}{c}{Northern Vietnamese} \\
\cmidrule(lr){4-5} \cmidrule(lr){6-7} \cmidrule(lr){8-9} \cmidrule(lr){10-11}
\textbf{Gen.} & \textbf{Label} & \textbf{Full Name} & \textbf{Term} & \textbf{Count} & \textbf{Term} & \textbf{Count} & \textbf{Term} & \textbf{Count} & \textbf{Term} & \textbf{Count} \\

\midrule
1st & MM & mother's mother & Grandmother & 3949 & Grootmoeder & 529 & Abuela & 8577.5 & bà ngoại & 210 \\ 

&   &   & Grandma & 882 & Oma & 548.5 & Yaya & 10 & bà & 411.5 \\

&   &   & Gran & 25.5 & &   &   &   &   &\\

\rowcolor[gray]{0.9}
& MF & mother's father & Grandfather & 3189 & Grootvader  & 720   & Abuelo & 8075.5 & ông ngoại & 125 \\

\rowcolor[gray]{0.9}
&   &   & Grandpa & 399.5 & Opa & 361.5 & Yayo & 2 & ông & 1210 \\

& FM & father's mother & Grandmother & 3949 & Grootmoeder & 529 & Abuela & 8577.5 & bà nội & 266 \\

&   &   & Grandma & 882 & Oma & 548.5 & Yaya & 10 & bà & 411.5 \\

&   &   & Gran & 25.5 & &   &   &   &   & \\

\rowcolor[gray]{0.9}
& FF & father's father & Grandfather & 3189 & Grootvader  & 720   & Abuelo & 8075.5 & ông nội   & 290 \\

\rowcolor[gray]{0.9}
&   &   & Grandpa & 399.5 & Opa & 361.5 & Yayo & 2 & ông & 1210 \\

\midrule

2nd & M  & mother & Mother & 65458 & Moeder & 18009 & Madre & 64464 & mẹ & 18004 \\

&    &        & Mom    & 29849 & Mama   & 1511  & Mama  & 1295  &    &       \\

&    &        & Mommy  & 707   & Ma     & 1186  & Mami  & 1071  &    &       \\

&    &        & Momma  & 388   &        &       & Mamá  & 52914 &    &       \\

\rowcolor[gray]{0.9}
& F  & father & Father & 63318 & Vader  & 19939 & Padre & 77214 & bố & 5021 \\
\rowcolor[gray]{0.9}
&    &        & Dad    & 31100 & Papa   & 896   & Papa  & 1250  &  cha  & 3390 \\
\rowcolor[gray]{0.9}
&    &        & Daddy  & 3017  & Pa     & 1346  & Papi  & 512   &    &       \\
\rowcolor[gray]{0.9}
&    &        &        &       &        &   & Papá  & 47494 &    &       \\

& MZy & mother's younger sister & Aunt   & 1022  & Tante  & 204.75 & Tía   & 1404.75 & dì & 243 \\

&     &                          & Auntie & 29    &        &        & Tita  & 1.5     & 
&     \\

\rowcolor[gray]{0.9}
& MBy & mother's younger brother & Uncle  & 1501.75 & Oom   & 509    & Tío   & 1404.75 & cậu & 237 \\
\rowcolor[gray]{0.9}
&     &                          &        &         &       &        & Tito  & 2.5     &  &     \\

& MZe & mother's elder sister    & Aunt   & 1022  & Tante  & 204.75 & Tía   & 1404.75 & bác gái & 2.5 \\
&     &                          & Auntie & 29    &        &        & Tita  & 1.5     & bác     & 96.5 \\

\rowcolor[gray]{0.9}
& MBe & mother's elder brother   & Uncle  & 1501.75 & Oom   & 509    & Tío   & 1404.75 & bác trai & 2 \\
\rowcolor[gray]{0.9}
&     &                          &        &         &       &        & Tito  & 2.5     & bác      & 96.5 \\

& FZy & father's younger sister  & Aunt   & 1022  & Tante  & 204.75 & Tía   & 1404.75 & cô       & 435 \\
&     &                          & Auntie & 29    &        &        & Tita  & 1.5     &          &     \\

\rowcolor[gray]{0.9}
& FBy & father's younger brother & Uncle  & 1501.75 & Oom   & 509    & Tío   & 1404.75 & chú      & 388 \\
\rowcolor[gray]{0.9}
&     &                          &        &         &       &        & Tito  & 2.5     &          &     \\

 & FZe & father's elder sister    & Aunt   & 1022  & Tante  & 204.75 & Tía   & 1404.75 & bác gái  & 2.5 \\
 &     &                          & Auntie & 29    &        &        & Tita  & 1.5     & bác      & 96.5 \\

\rowcolor[gray]{0.9}
 & FBe & father's elder brother   & Uncle  & 1501.75 & Oom   & 509    & Tío   & 1404.75 & bác trai & 2 \\
 \rowcolor[gray]{0.9}
 &     &                          &        &         &       &        & Tito  & 2.5     & bác      & 96.5 \\
 
\midrule

3rd & Zy & younger sister & Sister & 9587.5 & Zus & 2029.5 & Hermana 
 & 13610.5 & em & 2425.5 \\
 &    &                 & Sis    & 43.5   & Zusje & 826   & Tata    & 82      & em gái & 849 \\

\rowcolor[gray]{0.9}
& By & younger brother & Brother & 11687 & Broer & 2968 & Hermano & 22723 & em & 2425.5 \\
\rowcolor[gray]{0.9}
 &    &                  & Bro     & 63.5  & Broertje & 565 & Tete & 4 & em trai & 459 \\

 & Ze & elder sister & Sister & 9587.5 & Zus & 2029.5 & Hermana & 13610.5 & chị & 1752 \\
 &    &               & Sis    & 43.5   &       &       & Tata    & 82      &     &     \\

\rowcolor[gray]{0.9}
 & Be & elder brother & Brother & 11687 & Broer & 2968 & Hermano & 22723 & anh & 3173 \\
 \rowcolor[gray]{0.9}
 &    &                & Bro     & 63.5  &       &      & Tete    & 4       &     &     \\
 
\midrule

4th & D & daughter & Daugther & 23571 & Dochter & 10378 & Hija & 68744 & con & 5296.5 \\
    &   &          &          &        & Dochtertje & 1580 &       &       & con gái & 2543 \\

\rowcolor[gray]{0.9}
 & S & son & Son & 28815 & Zoon & 8737 & Hijo & 94575 & con & 5296.5 \\
 \rowcolor[gray]{0.9}
 &   &     &     &        & Zoontje & 3753 &      &       & con trai & 2370 \\

 & ZyD & younger sister's daugter & Niece & 326.25 & Nicht & 71.75 & Sobrina & 746 & cháu & 275.33 \\
 &     &                              &       &         & Nichtje & 121.75 &         &       & cháu họ & 4 \\

\rowcolor[gray]{0.9}
 & ZyS & yougher sister's son & Nephew & 339.25 & Neef & 177.75 & Sobrino & 1067 & cháu & 275.33 \\
 \rowcolor[gray]{0.9}
  &     &                         &        &         & Neefje & 76.5 &         &       & cháu họ & 4 \\

 & ByD & younger brother's daughter & Niece & 326.25 & Nicht & 71.75 & Sobrina & 746 & cháu & 275.33 \\
 &     &                             &       &         & Nichtje & 121.75 &         &       & cháu họ & 4 \\

\rowcolor[gray]{0.9}
 & ByS & younger brother's son & Nephew & 339.25 & Neef & 177.75 & Sobrino & 1067 & cháu & 275.33 \\
 \rowcolor[gray]{0.9}
 &     &                        &        &         & Neefje & 76.5 &         &       & cháu họ & 4 \\

 & ZeD & elder sister's daughter & Niece & 326.25 & Nicht & 71.75 & Sobrina & 746 & cháu & 275.33 \\
 &     &                          &       &         & Nichtje & 121.75 &         &       & cháu họ & 4 \\

\rowcolor[gray]{0.9}
 & ZeS & elder sister's son & Nephew & 339.25 & Neef & 177.75 & Sobrino & 1067 & cháu & 275.33 \\
 \rowcolor[gray]{0.9}
 &     &                      &        &         & Neefje & 76.5 &         &       & cháu họ & 4 \\

 & BeD & elder brother's daughter & Niece & 326.25 & Nicht & 71.75 & Sobrina & 746 & cháu & 275.33 \\
     &     &                          &       &         & Nichtje & 121.75 &         &       & cháu họ & 4 \\

\rowcolor[gray]{0.9}
 & BeS & elder brother's son & Nephew & 339.25 & Neef & 177.75 & Sobrino & 1067 & cháu & 275.33 \\
 \rowcolor[gray]{0.9}
     &     &                     &        &         & Neefje & 76.5 &         &       & cháu họ & 4 \\
     
\midrule

5th & DD & daughter's daughter & Granddaughter & 370 & Kleindochter & 87 & Nieta & 1304 & cháu & 275.33 \\
     &    &                     &               &     &              &    &       &      & cháu ngoại & 30 \\

\rowcolor[gray]{0.9}
 & DS & daughter's son & Grandson & 478 & Kleinzoon & 117 & Nieto & 1685.5 & cháu & 275.33 \\
 \rowcolor[gray]{0.9}
     &    &               &          &     &           &     &       &        & cháu ngoại & 30 \\

 & SD & son's daughter & Granddaughter & 370 & Kleindochter & 87 & Nieta & 1304 & cháu & 275.33 \\
     &    &               &               &     &              &    &       &      & cháu nội & 25 \\

\rowcolor[gray]{0.9}
 & SS & son's son & Grandson & 478 & Kleinzoon & 117 & Nieto & 1685.5 & cháu & 275.33 \\
 \rowcolor[gray]{0.9}
     &    &          &          &     &           &     &       &        & cháu nội & 25 \\
\bottomrule
\end{tabular}}
\end{table*}

\end{document}